# Rigid Point Registration with Expectation Conditional Maximization

Jing Wu

**Abstract**— This paper addresses the issue of matching rigid 3D object points with 2D image points through point registration based on maximum likelihood principle in computer simulated images. Perspective projection is necessary when transforming 3D coordinate into 2D. The problem then recasts into a missing data framework where unknown correspondences are handled via mixture models. Adopting the Expectation Conditional Maximization for Point Registration (ECMPR), two different rotation and translation optimization algorithms are compared in this paper. We analyze in detail the associated consequences in terms of estimation of the registration parameters theoretically and experimentally.

## 1 INTRODUCTION

REGISTRATION is the process of transforming different sets of data into the same coordinate system, with wide application in medical image analysis [1-3], human pose estimation [4], etc. Point registration (PR) is frequently met in image analysis and computer vision to find an optimal alignment between two sets of points. This problem can be separated by two processes, namely, 1) Find point-to-point correspondences and 2) estimate the transformation allowing the alignment of the two sets. Existing solvers to PR problem can be roughly divided into three categories: Iterative Closest Point (ICP) algorithm [5-6], soft assignment methods [7-8] and probabilistic methods [9-10].

ICP uses sampling processes, either deterministic or based on heuristics by outlier rejection. Although ICP is attractive for its efficiency, it can be easily trapped into local minima, which makes it sensitive to both initialization and the choice of a threshold for match acceptance. An alternative of ICP is soft assignments within a continuous optimization frame work. However, this algorithm needs optimal entries for assignment matrix, M, and satisfy the constraints on M, thus providing one-to-one assignments for inliers and many-to-one assignments for outliers. As a consequence, the convergence properties are not guaranteed in the presence of outliers.

In this paper, a probabilistic point registration is used with Gaussian mixture models (GMMs). The point-to-point assignment is to find the missing data with maximum likelihood. The algorithm of choice is the expectation conditional maximization (ECMPR) algorithm [9] to solve the problem of registering 3D data points on object with 2D data points on image plane. The challenge of this problem is to transform the 3D data into 2D by perspective projection. When the human eye looks at a scene, objects in the distance appear smaller than objects close by - this is known as perspective. While orthographic projection ignores this effect to allow accurate measurements, perspective definition shows distant objects as smaller to provide additional realism [11]. The difference between this paper and [9] is that there is inadequate observation data rather than redundant ones, thus there are no outlier points in this paper.

This paper has the following contributions:
- Perspective projection is used to transform 3D data points into 2D image plane. The camera's position, orientation, and field of view control the behavior of the projection transformation. Thus, four transformation matrices, namely, scaling, perspective projection, rotation, and translation, should be taken into consideration.
- Two different algorithms for rotation matrix and translation vector optimization are used in this paper. The first algorithm is traversal scan that searches the whole angle and distance space to get the optimal solution, which is accurate but time consuming. The second algorithm is based on least-squares estimation [10], which is efficient but is less accurate. The performance of these two algorithms is compared by experimental results in Section 3.
- The impact of initial conditions, model point radius, added Gaussian noise and the use of anisotropic covariance and isotropic ones are discussed in detail in the results part of the paper.

The remainder of this paper is organized as follows: In Section 2, the PR problem is formed and ECMPR algorithm is listed for rigid point sets registration. In Section 3 shows the experimental results with testing data.. And Section 4 is the conclusions.

## 2 PROBLEM FORMULATION

### 2.1. Mathematical Notations

Two data sets, namely 3D brace points and 2D image points, will be considered in this paper. We denote $\mathcal{Y} = \{Y_j\}_{1 \leq j \leq m}$ the 2D coordinates of a set of observed image data points and $\mathcal{X} = \{X_i\}_{1 \leq i \leq n}$ the 3D coordinates of a set of model points of brace bead. The object is to find correspondence of observed data $\mathcal{Y}$ to model data $\mathcal{X}$. The correspondence is denoted by $\mathcal{Z}$ as missing data where $\mathcal{Z}_j = i$ means the j$^{th}$ point in $\mathcal{Y}$ is corresponding to i$^{th}$ point in $\mathcal{X}$. The 3D model points lie in global coordinate as shown is **Fig 1**; the 2D image points on the image plane in image coordinate; while the camera has its camera coordinate, as shown in **Fig. 2**. Hence, a coordinate transformation is necessary to compare points in the same coordinate system. The image coordinate can be easily transformed to the camera coosdrdinate by moving the origin of image to that of the camera in xy plane while its z axis is assigned as focal distance in camera coordinate. The next task is to register global coordinate to the camera coordinate. In this paper, a perspective projection is considered [12]. Perspective projection is to show distant objects as smaller to provide additional realism. The perspective projection is a $4 \times 4$ matrix built from four component homogenous matrices, PSRT where P is perspective matrix that is responsible for fore-shortening; S is scale matrix that adjusts for aspect ratio; R is rotation matrix that determines which direction to look at and T is translation matrix that determines where the camera is. The coordinate transformation is as follows

$$x' = PSRTx \quad (1)$$

$$= \begin{pmatrix} f & 0 & 0 & 0 \\ 0 & f & 0 & 0 \\ 0 & 0 & f & 0 \\ 0 & 0 & 1 & 1 \end{pmatrix} \begin{pmatrix} s & 0 & 0 & 0 \\ 0 & s & 0 & 0 \\ 0 & 0 & s & 0 \\ 0 & 0 & 0 & 1 \end{pmatrix} \begin{pmatrix} r_{11} & r_{12} & r_{12} & 0 \\ r_{21} & r_{22} & r_{23} & 0 \\ r_{31} & r_{32} & r_{33} & 0 \\ 0 & 0 & 0 & 1 \end{pmatrix} \begin{pmatrix} 1 & 0 & 0 & t_1 \\ 0 & 1 & 0 & t_2 \\ 0 & 0 & 1 & t_3 \\ 0 & 0 & 0 & 1 \end{pmatrix} \begin{pmatrix} x_1 \\ x_2 \\ x_3 \\ 1 \end{pmatrix}$$

where x is the 3D points $(x_1, x_2, x_3)$ that is converted to 4x1 matrix by adding a one (homogeneous points) at the end of the vector. f is focal distance (f=40 inch), s is scale coefficient (s=1.017mm/pixel), $r_{ij}$ is elements of rotation matrix while $t_1$, $t_2$, $t_3$ are translation in x, y and z axis respectively.

Since f, s are constant, equation (1) can be converted to

$$\mu(x_i; \Theta) = Rx_i + t$$
$$\mu'(x_i; \Theta) = \frac{fs\mu(x_i; \Theta)}{\mu(3,1)} \quad (2)$$

where $\Theta := \{R, t\}$, R is $3 \times 3$ rotation matrix and t is $3 \times 1$ translation vector. $\mu: \mathbb{R}^3 \to \mathbb{R}^2$ is a mapping from 3D coordinate to 2D. We will refer to the parameter vector $\Theta$ as the registration parameters and a superscripted $\Theta^*$ denotes the optimized value of that parameter. The distance used in this paper is Mahalanobis distance, denoted by $\|X - Y\|_{\Sigma}^2 = (X - Y)^T \Sigma^{-1} (X - Y)$ where $\Sigma$ is a $3 \times 3$ symmetric positive definite matrix.

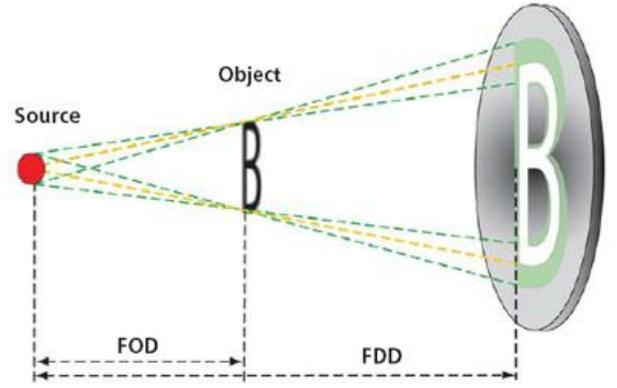

Fig. 2 Perspective projection

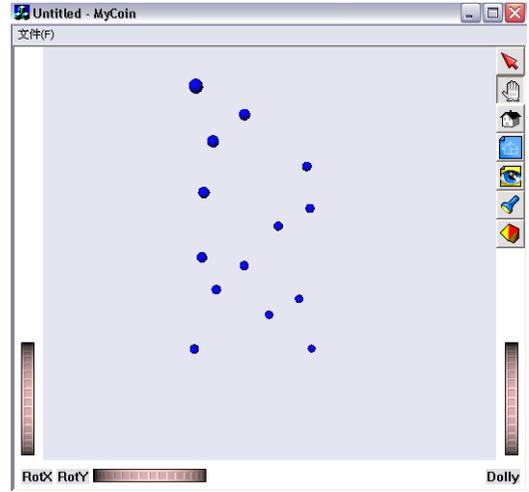

Fig.1 3D data of bead on the brace

### 2.2. Expectation conditional maximization for point Registration

In this paper, we will optimize the parameters by expectation conditional maximization such that the center of observed data is constrained to coincide with the transformed model points $\mu'(x_i; \Theta)$. Suppose $\mathcal{X}$ and $\mathcal{Y}$ follows Gaussian distribution, the expected observed data log likelihood is a function of both the registration parameters and the covariance matrices:

$$\mathcal{E}(\Theta, \Sigma_1 ... \Sigma_n | \mathbf{Y}, \mathbf{Z}) = E_Z[\log P(\mathbf{Y}, \mathbf{Z}; \Theta, \Sigma_1, ... \Sigma_n) | \mathbf{Y}] \quad (3)$$

A powerful method for finding maximal likelihood solutions in the presence of hidden variables is to replace the observed-data log-likelihood with the complete-data log-likelihood and to maximize the expected complete-data likelihood conditioned by the observed data. The criterion to be maximized becomes

$$\mathcal{E}(\Theta, \Sigma_1 ... \Sigma_n | \mathbf{Y}, \mathbf{Z}) = \sum_Z P(\mathbf{Z} | \mathbf{Y}, \Theta, \Sigma)$$
$$= -\frac{1}{2} \sum_{j=1}^{m} \sum_{i=1}^{n} \alpha_{ji}^q \left( \|\mathbf{Y}_j - \mu'(\mathbf{X}_i, \Theta)\|_{\Sigma_i^q}^2 + \log |\Sigma_i| \right) \quad (4)$$

Where $\alpha_{ji}^q$ is posterior probability described by

$$\alpha_{ji} = \frac{|\Sigma_i|^{-\frac{1}{2}} \exp\left(-\frac{1}{2}\|\mathbf{Y}_j - \boldsymbol{\mu}'(\mathbf{X}_i;\Theta)\|^2_{\Sigma_i}\right)}{\sum_{k=1}^{n} |\Sigma_i|^{-\frac{1}{2}} \exp\left(-\frac{1}{2}\|\mathbf{Y}_j - \boldsymbol{\mu}'(\mathbf{X}_i;\Theta)\|^2_{\Sigma_i}\right) + \Phi_{3D}} \quad (5)$$

where $\Phi_{3D}$ corresponds to the outlier component in the case of 3D point registration: $\Phi_{3D} = 1.5\sqrt{2\pi}r^{-3}$

Thus the parameters to be optimized are

$$\Theta^{q+1} = \arg\min_{\Theta} \frac{1}{2} \sum_{j=1}^{m}\sum_{i=1}^{n} \alpha_{ji}^q \|\mathbf{Y}_j - \boldsymbol{\mu}'(\mathbf{X}_i,\Theta)\|^2_{\Sigma_i^q} \quad (6)$$

And

$$\Sigma^{q+1} = \frac{\sum_{j=1}^{m}\sum_{i=1}^{n} \alpha_{ji}^q (\mathbf{Y}_j - \boldsymbol{\mu}'(\mathbf{X}_i,\Theta))(\mathbf{Y}_j - \boldsymbol{\mu}'(\mathbf{X}_i,\Theta))^T}{\sum_{j=1}^{m}\sum_{i=1}^{n} \alpha_{ji}^q} \quad (7)$$

With the virtual observation $W_i$ and its weight $\lambda_i$:

$$\mathbf{W}_i = \frac{1}{\lambda_i} \sum_{j=1}^{m} \alpha_{ji} \mathbf{Y}_j \quad (8)$$

$$\lambda_i = \sum_{j=1}^{m} \alpha_{ji} \quad (9)$$

### 2.3. The ECMPR-rigid algorithm

1. *Initialization*: Set $\mathbf{R}^q = \mathbf{I}, \mathbf{t}^q = \mathbf{0}$. Choose the initial covariance matrices $\Sigma^q = \mathbf{I}$,

2. *Coordinate Transformation*: project the $X_i$ into the camera coordinate by rotation, translation, perspective projection, and scaling by (2)

3. *E-Step*: Evaluate the posteriors $\alpha_{ji}^q$ from (5), $\lambda_i^q$ from (8), $\mathbf{W}_i^q$ from (7) using the current parameters $\mathbf{R}^q, \mathbf{t}^q$ and $\Sigma_i^q$. When calculating $\mathbf{W}_i^q$, an infinity check is necessary because the number of points in $\mathbf{Y}_j$ is less than that in $\mathbf{X}_i$. This leads to infinity in the columns of $\mathbf{W}_i^q$ which dose not have a corresponding point in $\mathbf{Y}_j$ and $\lambda_i = 0$ for the points that have no correspondence to the observed points.

4. *CM-steps*: (two different searching methods are used in this paper, thus they are listed separately in a and b)
   a *Traversal:*
      i. Set the object function as
      $$J = \frac{1}{2}\sum_{j=1}^{m}\sum_{i=1}^{n} \alpha_{ji}^q \|\mathbf{Y}_j - \boldsymbol{\mu}'(\mathbf{X}_i,\Theta)\|^2_{\Sigma_i^q}$$
      Search optimal $\mathbf{R}^*$ to minimize J by using the current $\mathbf{t}^q$ in the range of $[0, 2\pi]$.
      ii. Search optimal $\mathbf{t}^*$ to minimized J by using the updated $\mathbf{R}^{q+1}$ in the range of [-1000,1000].
      iii. Estimate the new covariances from (7) with the current posteriors, the new rotation matrix and the new translation vector.

   b *Least-Squares Estimation [10]:*
      i. Calculate the correspond between $X_i$ and $Y_j$ using the current posterior and re-arrange $Y_j$ accordingly
      ii. Calculate the SVD of $\Sigma_{xy} = \mathbf{UDV}^T$ and
      $$\mathbf{S} = \begin{cases} \mathbf{I} & \text{if } \det(\Sigma_{xy}) \geq 0 \\ \text{diag}(1,1,...,1,-1) & \text{if } \det(\Sigma_{xy}) < 0 \end{cases}$$
      
      $$\boldsymbol{\mu}_x = \frac{1}{n}\sum_{i=1}^{n}\mathbf{x}_i$$
      
      Where $\boldsymbol{\mu}_y = \frac{1}{n}\sum_{i=1}^{n}\mathbf{y}_i$ and n
      
      $$\sigma_x^2 = \frac{1}{n}\sum_{i=1}^{n}\|\mathbf{x}_i - \boldsymbol{\mu}_x\|^2$$
      
      $$\sigma_y^2 = \frac{1}{n}\sum_{i=1}^{n}\|\mathbf{y}_i - \boldsymbol{\mu}_y\|^2$$
      
      $$\Sigma_{xy} = \frac{1}{n}\sum_{i=1}^{n}(\mathbf{y}_i - \boldsymbol{\mu}_y)(\mathbf{x}_i - \boldsymbol{\mu}_x)^T$$
      
      is the number of points
      
      iii. When rank $(\Sigma_{xy}) \geq m-1$ (m is the dimension of the data, and here m=3), the optimal transformation parameters are determined as follows
      $$\mathbf{R} = \mathbf{USV}^T \quad (10)$$
      $$\mathbf{t} = \boldsymbol{\mu}_y - c\mathbf{R}\boldsymbol{\mu}_x$$
      
      iv. Estimate the new covariance matrix ($\Sigma^{q+1}$) from (7) with the current posteriors, the new rotation matrix and the new translation vector.

5. *Convergence:* Compare the new and current rotations. If $\|R^{q+1} - R^q\|^2 < 10^{-6}$, then go to the Classification step. Else, set the current parameter values to their new values and return to the Step 2.

6. *Classification:* Assign each observation to a model point based on the maximum a posteriori (MAP) principle:
$$z_j = \arg\max_i \alpha_{ji}^q$$

## 3 EXPERIMENTAL RESULTS

We carried out several experiments with ECMPR-rigid algorithm on the platform of Matlab 7.11.0 (R 2010b). First, two different R, t searching algorithm was applied to testing data to assess the performance of the method as summarized in **Table I and Fig.3**. Then, real data point from both anteroposterior and lateral X-ray images were calculated by the ECMPR-rigid algorithm. The results are shown in **Table II-IV and Fig.4-7**.

For the testing data, we considered 14 model points, corresponding to the clusters' centers in the mixture model, as well as 14 observations generated from the model points by, perspective projection, scaling, rotation and translation and corrupted by noise. The model points are rotated by -20, 20 10 degrees in x, y, z axis respectively and then translated by a vector [100, -400, 200]. The projection factor is f=1016 mm and scaling factor is s=0.175 mm/pixel. We simulated both noise-free cases in the second and forth row and Gaussian noise case in the first and third row. This noise is centered at each model point and is drawn from 1D Gaussian probability distributions with variance of 1 along each dimension. The initial parameters are rotation angle as identity matrix and translation vector as null vector.

Two different optimization algorithm for R, t searching was studied, namely, traversal and least-squares estimation(LSE). The percentage of correct matches is defined by the number of observations that were correctly classified over the total number of observations. This classification is based on the MAP principle: each observation $j$ is assigned to the cluster $i$ such that $j = \arg\max_i(\alpha_{ji})$. It can be shown from Table 1 that LSE uses much less time than the traversal algorithm. Moreover, LSE needs less iterate to reach a much smaller convergence threshold than the traversal algorithm. The matching errors are the same for both algorithms. But the traversal algorithm uses big the searching step it will have larger matching error than the LSE. The shortcoming of LSE, however, is its accuracy.

Additionally, we performed a large number of trials with ECMPR-rigid traversal algorithm in the anisotropic covariance case (first row in Table 1). The model points are rotated with an angle that varies between 0 and 180 degrees. **Fig. 3** show the percentage of correct matches, the relative error in rotation, and the relative error in translation as a function of the ground-truth rotation angle between the sets of data and the model points. The plotted curves correspond to the mean values and the variances computed over each rotation.

TABLE I
COMPARISON BETWEEN TWO OPTIMIZATION ALGORITHM

| Algorithm | Simulated noise | Covariance model | Iteration number |
|---|---|---|---|
| Traversal | anisotropic | anisotropic | 10 |
| Traversal | - | anisotropic | 2 |
| Least-Squares Estimation | anisotropic | anisotropic | 2 |
| Least-Squares Estimation | - | anisotropic | 2 |

| Algorithm | Error in rotation | Process time | Correct match% |
|---|---|---|---|
| Traversal | $10^{-6}$ | 49.374923s | 100% |
| Traversal | $10^{-6}$ | 47.584925s | 100% |
| Least-Squares Estimation | $10^{-10}$ | 0.054400s | 100% |
| Least-Squares Estimation | $10^{-10}$ | 0.292593s | 100% |

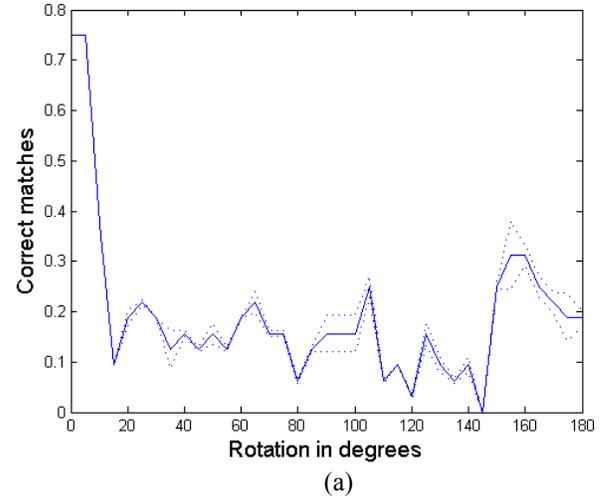

(a)

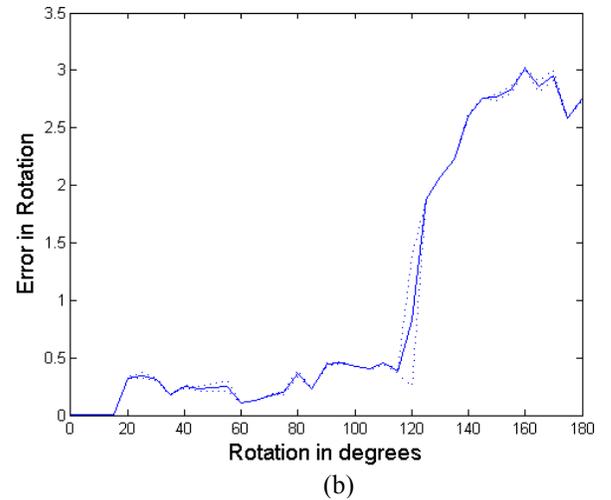

(b)

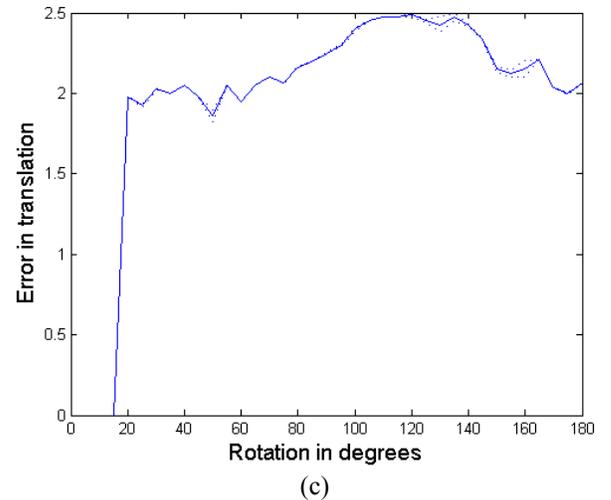

(c)

Fig. 3 Statics obtained with ECMPR-rigid over four trials. The percentage of correct matches and relative errors in rotation and translation are shown as a function of the ground-truth rotation angle between the set of data points and the set of model points. The three curves correspond to the means (central curves) and to the means +/- the standard deviation (upper and lower cures) computed over 4 trials (a) Correct matches. (b) Error in rotation. (c) Error in translation.

## 4 Conclusions

In this paper, we address the problem of perspective projection from 3D coordinate to 2D coordinate and matching rigid shapes through robust point registration. The proposed approach uses maximum likelihood with hidden variables for model based clustering. We used variant EM algorithm- E-CM, to maximize the expected complete-data log-likelihood and preserving the convergence properties of EM. We compared two different algorithms namely, traversal and least-squares estimation, for rotation matrix and translation vector optimization. Experimental results show robust of the proposed algorithm b adding Gaussian noise in conjunction with point registration. By comparing the two R, t optimization algorithms, the traversal algorithm shows accuracy while lack of efficiency and least-squares estimation is vice-versa. In the future, application of ECMPR to medical image registration [2-3], HCI in mobile device [13-14], human pose estimation [4] will be explored.